# Devnagari Handwritten Numeral Recognition using Geometric Features and Statistical combination Classifier

Vikas J. Dongre
Department of Electronics & Telecommunication
Government Polytechnic, Gondia (MS)
India
Email: dongrevj@yahoo.co.in

Vijay H. Mankar
Department of Electronics & Telecommunication
Government Polytechnic, Nagpur (MS)
India
Email: vhmankar@gmail.com

*Abstract* — This paper presents a Devnagari Numerical recognition method based on statistical discriminant functions. 17 geometric features based on pixel connectivity, lines, line directions, holes, image area, perimeter, eccentricity, solidity, orientation etc. are used for representing the numerals. Five discriminant functions viz. Linear, Quadratic, Diaglinear, Diagquadratic and Mahalanobis distance are used for classification. 1500 handwritten numerals are used for training. Another 1500 handwritten numerals are used for testing. Experimental results show that Linear, Quadratic and Mahalanobis discriminant functions provide better results. Results of these three Discriminants are fed to a majority voting type Combination classifier. It is found that Combination classifier offers better results over individual classifiers.

I.  INTRODUCTION

Machine simulation of human functions has been a challenging research field since the advent of digital computers. In some areas which require certain amount of intelligence such as number crunching or chess playing, tremendous improvements are achieved. On the other hand, humans still outperform even the most powerful computers in the relatively routine functions such as vision. Imitating the human abilities in computers is not an easy task due to high context sensitivity. Machine simulation of human reading is one of these areas which have been the field of intensive research from early days of computer. Yet it is still far from the final frontier. Optical Character recognition (OCR) and document processing has become the need of time with the popularization of desk top publishing and usage of internet. OCR also finds application in postal mail sorting, Automatic Bank cheque clearance, Blind person aid, automatic form processing. OCR involves recognition of characters from digitized images of optically scanned document pages. The characters thus recognized from document pages are coded with American Standard Code for Information Interchange (ASCII) or some other standard codes like UNICODE for storing in a file which can be further edited as any other file created with some word processing software or editor.

A lot of research has been done in developed countries for English, European and Chinese languages. But still there is a dearth of need to carry out research in Indian languages. Most of the Indian scripts are distinguished by the presence of matras (character modifiers) and Shirorekha (line on the top of the characters to form a word) in addition to main characters as against the English script that has no matras or Shirorekha. Therefore, algorithms developed for English language are not directly applicable to Indian scripts. Devnagari is the most popular script in India. Hindi, the national language of India, is written in the Devnagari script and is spoken by more than 500 million people. Moreover, Hindi is the third most popular language in the world. Devnagari script is also used for writing Marathi, Sanskrit, Konkani and Nepali languages. Many OCRs for Devnagari script have been reported. However, none of these has attempted the handwritten Devnagari script consisting of composite characters that involve both the main characters and matras. In this work we are specifically considering Devnagari numeral recognition. Sample images of Devnagari numerals from 0 to 9 are shown in fig. 1.

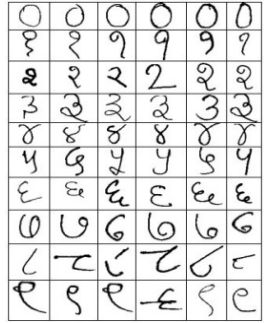

Figure 1: Sample Data.

Remarkable work in Devnagari Optical Character Recognition (DOCR) is carried out by R. M. K. Sinha and V. Bansal [1-2]. A general Review of Pattern Recognition can be found in [3-6]. Various feature extraction methods as well as classifiers are proposed in the literature for classification of handwritten numerals and characters. Feature extraction methods include chain code, Gabor filter, Hough transformations, moments, structural features, principal component analysis. Classifiers include minimum mean distance, k-nearest neighbor techniques (KNN), support vector machines (SVM), neural Networks (NN), fuzzy based approaches and their combinations. A study of different pattern recognition methods are given in [4, 7, 8]. Lot of research work in English numeral recognition is reported in journals [9]. But recognition of handwritten numerals for Indian scripts is still a challenging task.

P.M. Patil et al. [10] used Rotation scale and translation invariant moments for handwritten Devanagari numeral recognition. A method based on Hu moments for the recognition of handwritten Devnagari numerals is proposed by Ramteke et al. [11]. They used seven central invariant moments as features. V. Vijaya Kumar et al. [12] presented mathematical morphology based method for classification and recognition of English handwritten digits. M. Hanmandlu et al. [13] presented Fuzzy model based recognition of handwritten English and Devnagari numerals. G. G. Rajput et al. [14] presented a method for Isolated Marathi Numeral Recognition using Density and Central Moment Features. Works reported in [15-17] generated comprehensive Devnagari numeral and character databases.

We will now discuss some of our earlier work towards in the character recognition. We segmented the Devnagari documents Using Histogram Approach [17]. It was concluded that segmentation of compound characters in Devnagari is a challenging job. The same algorithm was modified for segmenting the scanned images. Database of 5137 numerals and 20305 characters is generated [16]. The self-generated numeral database is used in this work for numeral recognition.

Rest of the paper is organized as follows. Section 2 describes the process overview. Section 3 describes various preprocessing techniques. Section 4 deals with the feature extraction. In Section 5, the recognition strategy is presented. In Section 6, design of three input and five input combination classifier is described. Section 7 briefly reports the results of Devnagari Numeral recognition. Conclusions and scope of future work is discussed in section 8.

II. PROCESS OVERVIEW

A self-generated numeral database [16] is used for both training and testing phase of the work. Data was collected from 750 persons of different age, gender, background, qualification, locality and profession by providing them a standard plain paper sheet as shown in fig. 2. It was digitized, segmented and stored in 10 folders. Each folder contains set of scanned numeral images. The database is divided into two disjoint sets: one set of 150 images per numeral for training and another set of 150 images per numeral for testing. All the images are preprocessed in various steps for normalization, extracting geometric and structural features from them. Discernment functions viz. Linear, Quadratic, Diaglinear, Diagquadratic and Mahalanobis are used for classification of numeral classification. Finally the result of three most efficient discriminators viz. Linear, Quadratic and Mahalanobis are fed to combination classifier in view to improve the overall accuracy. Overview of the process is shown in figure 3. The steps are mentioned below.

(i) Image acquisition and segmentation
(ii) Binarization of image

Figure 2: Data collection sample sheet.

(iii) Image inversion
(iv) Image cleaning and Noise removal
(v) Image Discoursing
(vi) Image normalization
(vii) Image thinning
(viii) Feature extraction
(ix) Recognition

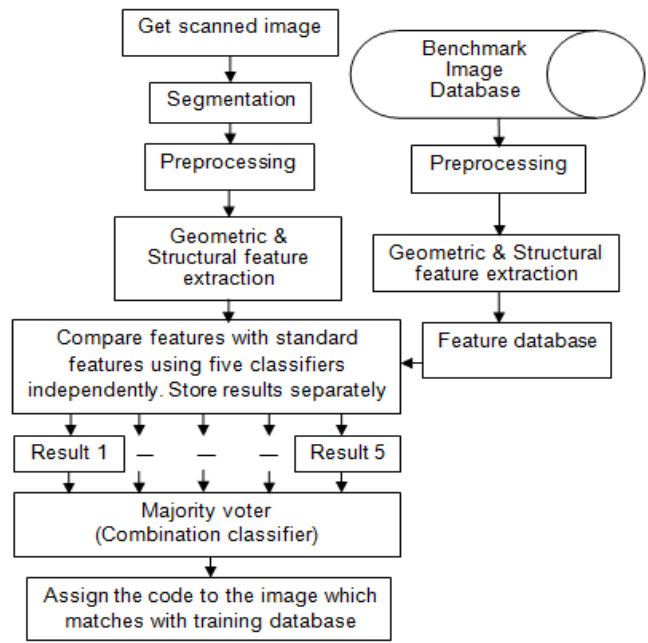

Figure 3: Numeral recognition process.

III. PREPROCESSING

A. *Image acquisition and segmentation*

The paper sheet having data written by 750 writers (fig. 2) is digitized scanner at 300 dpi. Various program modules were developed for segmentation of images. Database of 5137 numeral images is stored using .tiff format in ten folders. Sample database is shown in fig.1. In this work, 3000 image are used after preprocessing.

B. *Binarization*

Binarization is the first step in preprocessing. In pattern recognition we are interested only in the shape and contour of the image. Color or gray levels of image are redundant. Binarization also reduces data storage space and further processing time. Image binarization converts an image of up to 256 gray levels into a two-tone image represented by 0 and 1 (fig. 4a).

C. *Image inversion*

Image is generally drawn by dark shade on white paper. We are interested in dark marks on the scanned images. Image inversion is needed to make actual written data pixels as 1 and remaining pixels as 0 for convenient processing (fig. 4b).

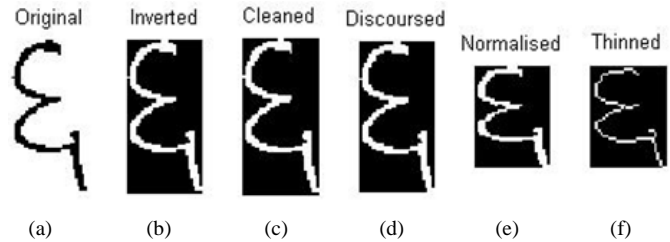

Figure 4: Preprocessing of Devnagari Numerals.

*D. Image cleaning and noise removal*

The scanned image may contain spurious marks or isolated dots. These isolated marks are removed to minimize error in further processing. Image is smoothed using dilation and erosion process so that connectivity of the strokes and curves may be properly maintained and unfilled regions on the strokes be filled.

*E. Image Discoursing*

Digitized image may contain some extra region having no information. A bound box is formed to fit the image in smallest rectangle (fig. 4d). This is used as input for normalization of image.

*F. Image normalization*

As the features to be extracted are geometry and structure based, all the images must be of standard dimension for meaningful comparison of the features. Hence the images are normalized. All the images are reshaped to have dimension 40 x 30 pixel. This is done by observing the standard shapes of Devnagari numerals (fig. 4e).

*G. Image Thinning*

Image thinning is done to minimize processing time as well as to find the predominant strokes in the image. Various filter operators are used to find number of strokes and their area can only be calculated on thinned image. Otsu's method is used for thinning in this work (fig. 4f).

After performing above preprocessing, the image is ready for extracting various geometric and structural features.

## IV. FEATURE EXTRACTION

Feature extraction and selection can be defined as extracting the most representative information from the raw data, which minimizes the within class pattern variability while enhancing the between class pattern variability [8]. In this work following 17 geometric and structural features are extracted from each image after normalization and thinning.

1. Number of horizontal lines
2. Number of vertical lines
3. Number of Right diagonal lines
4. Number of Left diagonal lines
5. Length of all horizontal lines
6. Length of all vertical lines
7. Length of all right diagonal lines
8. Length of all left diagonal lines
9. Euler Number
10. Convex Area
11. Filled Area
12. Solidity
13. Perimeter
14. Area
15. Eccentricity
16. Extent
17. Orientation

The structural features 1 to 8 are extracted using the algorithm by Dinesh Dileep [19]. As the name of feature implies, lines of different directivity, their count and sum of length of all lines is calculated and considered as structural features.

Geometric features from 9 to 17 are extracted using standard functions available in Matlab (20). They are explained below.

1. *Euler Number*: Scalar that specifies the number of objects in the region minus the number of holes in those objects.

2. *Convex Area*: Scalar that specifies the number of pixels in Convex Image.

3. *Filled Area*: Scalar specifying the number of on pixels in Filled Image.

4. *Solidity*: Scalar specifying the proportion of the pixels in the convex hull that are also in the region. Computed as Area/Convex Area.

5. *Perimeter*: It is the distance around the boundary of the region.

6. *Area*: It is the actual number of pixels in the region.

7. *Eccentricity*: It is defined as the eccentricity of the smallest ellipse that fits the skeleton of the image.

8. *Extent*: Scalar that specifies the ratio of pixels in the region to pixels in the total bounding box, computed as the Area divided by the area of the bounding box

9. *Orientation*: It is the angle (in degrees ranging from -90 to 90 degrees) between the x-axis and the major axis of the ellipse that has the same second-moments as the region.

150 images are used for each Devnagari numeral from 0 to 9. Hence 1500 images are used for training purpose. (Fig. 1 shows some of the collected numerals with stroke variations.). Thus a vector of 1500 x 17 is obtained. Another 1500 images are used for Devnagari numeral from 0 to 9 for testing purpose. A target matrix of 1500 x 1 was created as a reference for training and estimating.

V. NUMERAL RECOGNITION

In this work Discriminant analysis is used for classification of Devnagari numerals. Discriminant analysis uses training data to estimate the parameters of Discriminant functions of the predictor variables. Discriminant functions determine boundaries in predictor space between various classes. The resulting classifier discriminates among the classes (the categorical levels of the response) based on the predictor data [20]. Five discriminant functions viz. Linear, Quadratic, Diaglinear, Diagquadratic and Mahalanobis distance are used for classification independently.

a. *Linear*: Fits a multivariate normal density to each group with a pooled estimate of covariance.
b. *Diaglinear*: Similar to linear but with a diagonal covariance matrix estimate (naive Bayes classifiers).
c. *Quadratic*: Fits multivariate normal densities with covariance estimates stratified by group.
d. *Diagquadratic*: Similar to quadratic but with a diagonal covariance matrix estimate (naive Bayes classifiers).
e. *Mahalanobis*: Uses Mahalanobis distances with stratified covariance estimates.

As explained in previous section, 17 features of 1500 benchmark images (150 images for each numeral 0 to 9) are stored in *training* matrix (1500x17). Benchmark result *group* matrix (1500 x 1) is generated corresponding to training matrix. 17 Features of Test image to be recognized is stored in *sample* matrix. The specified algorithm classifies the test features and assigns appropriate *class* (0 to 9) to the test sample as per the *type* of discriminant function mentioned above. The syntax is as follows [20].

class = classify (sample, training, group, type)

In this work two sets of 150 images for each numerical were used. In first case training and sample data was kept identical. In second case training and sample (test) data was disjoint. All the discriminant types were tested for both the cases. A combination classifier is also designed. Individual results are fed to combination classifier. Individual and combination results are given in table I.

VI. COMBINATION CLASSIFIER

Various classification methods have their own superiorities and weaknesses. Different classifiers trained on the same data may not only differ in their global performances, but they also may show strong local differences. Each classifier may have its own region in the feature space where it performs the best [8]. Hence many times multiple classifiers are combined together to improve classification accuracy. Various schemes for combining multiple classifiers can be grouped into three main categories according to their architecture: 1) parallel, 2) cascading (or serial combination) and 3) hierarchical (tree-like). Here we have used a parallel combiner. Design procedure for this combiner is discussed below.

A. *Design of Combiner.*

In this study we have designed two combiners viz. Majority of three, majority of five.

With n inputs we can have majority combinations as follows.

i. All n are same.
ii. $^nC_r$ combinations of r majority.
iii. $^nC_{r-1}$ combinations of r-1 majority.
iv. $^nC_{r-2}$ combinations of r-2 majority.
v. It goes down till r-(r-1) majority i.e. 1. This indicates no two results are similar.

Also, $^nC_r = \dfrac{n!}{r!(n-r)!}$

*1. Majority of three Combiner*

In this case n=3 and r=3, 2, 1.

∴ $^nC_r$ i.e. $^3C_3 = 1$; $^3C_2 = 3$; $^3C_1 = 3$.

For three input majority detector, it is enough to check if two or more inputs are similar output. We can have maximum 3 combinations.

*Algorithm:*

i. Input the results of three classifiers to combiner.
ii. if i(1) = i(2), output= i(1).
iii. else if i(1) = i(3), output= i(1).
iv. else if i(2) = i(3), output= i(2).
v. else output= i(2) (assuming that all inputs are different as second classifier gives better result in our case)

In case all the results are different for individual classifiers, highest priority is assigned to second classifier. This is because in our work the second classifier i.e. Quadratic classifier offers best classification.

*2. Majority of Five Combiner*

In this case n=5 and r=5, 4, 3, 2, 1.

∴ $^nC_r$ i.e. $^5C_5 = 1$; $^5C_4 = 5$; $^5C_3 = 10$; $^5C_2 = 10$; $^5C_1 = 5$.

For five input majority detector, it is needed to check the following.

i. Check if any three inputs are similar. If yes, it will be taken as output of combiner. We can have maximum 10 combinations.
ii. If no, check if any two inputs are similar. We can have maximum 10 combinations.
iii. If no two inputs are similar output, i(2), i.e. Output of quadratic classifier is taken as output of combination classifier as it offers best classification.

*Algorithm:*

i. Input the results of five classifiers to combiner.
ii. if i(1) = i(2) = i(3) or i(1) = i(2) = i(4) or i(1) = i(2) = i(5) or i(1) = i(3) = i(4) or i(1) = i(3) = i(5) or i(1) = i(4) = i(5) ; output= i(1).
iii. else if i(2) = i(3) = i(4) or i(2) = i(3) = i(5) or i(2) = i(4) = i(5); output = i(2).
iv. else if i(3) = i(4) = i(5); output = i(3).
v. else if i(1) = i(2) or i(1) = i(3) or i(1) = i(4) or i(1) = i(5); output = i(1).
vi. else if i(2) = i(3) or i(2) = i(4) or i(2) = i(5) output = i(2).
vii. else if i(3) = i(4) or i(3) = i(5); output = i(3).
viii. else if i(4) = i(5); output = i(4).
ix. else output = i(2) (assuming that all inputs are different as second classifier gives better result in our case) .

VII. RESULTS

It can be seen that Linear, Quadratic and Mahalanobis discriminant functions outperforms Diaglinear and Diagquadratic discriminant functions individually. Results are degraded in second case as both training and test set are disjoint. Initially Individual results of all the discriminant function are fed as input to best of five combination classifier. It is found that Combination classifier performance is degraded in comparison with quadratic classifier which offers best classification results. This is obvious as the contribution of Diaglinear and Diagquadratic discriminant functions brings down the performance. Finally, Individual results of only Linear, Quadratic and Mahalanobis discriminant function are fed as input to combination classifier as these discriminant functions offer better classification. It is found that best of three combination classifier offers best results. It can also be seen that results for numeral 8 are highest (95.33% and 93.33% in case 1 and 2 respectively) (Table I). The exact result of Combination classifier for numerals 0 to 9 are plotted for training and test cases in fig. 5 and 6 respectively.

TABLE I: COMPARATIVE RESULTS FOR VARIOUS DISCRIMINANT FUNCTIONS.

| Discrimator function | % accuracy results for *Training* data on Devnagari Handwritten Numerals | | | | | | | | | | Avg. % Acuracy |
|---|---|---|---|---|---|---|---|---|---|---|---|
| | 0 | १ | २ | ३ | ४ | ५ | ६ | ७ | ८ | ९ | |
| Linear (L) | 85.33 | 87.33 | 55.33 | 54.67 | 72.67 | 79.33 | 70.00 | 74.67 | 90.67 | 54.67 | 72.47 |
| Quadratic (Q) | 94.00 | 84.67 | 74.00 | 54.67 | 86.67 | 75.33 | 90.00 | 86.00 | 95.33 | 66.00 | 80.67 |
| DiagLinear (DL) | 85.33 | 78.67 | 40.67 | 44.67 | 70.00 | 67.33 | 43.33 | 66.67 | 78.00 | 48.67 | 62.33 |
| DiagQuadratic (DQ) | 94.67 | 75.33 | 43.33 | 32.00 | 72.00 | 42.67 | 70.00 | 67.33 | 92.00 | 42.67 | 63.20 |
| Mahalanobis (M) | 93.33 | 58.67 | 78.00 | 83.33 | 84.67 | 82.00 | 75.33 | 83.33 | 82.00 | 82.00 | 80.27 |
| L+D+DL+DQ+M majority | 93.33 | 82.67 | 68.67 | 62.00 | 82.67 | 80.67 | 79.33 | 80.00 | 92.00 | 67.33 | 78.87 |
| **L+D+M majority** | 94.00 | 82.00 | 75.33 | 67.33 | 86.00 | 79.33 | 83.33 | 84.67 | 94.00 | 70.67 | **81.67** |

| Discrimator function | % accuracy results for *Test* data on Devnagari Handwritten Numerals | | | | | | | | | | Avg. % Acuracy |
|---|---|---|---|---|---|---|---|---|---|---|---|
| | 0 | १ | २ | ३ | ४ | ५ | ६ | ७ | ८ | ९ | |
| Linear (L) | 82.00 | 59.33 | 30.00 | 68.67 | 68.67 | 74.00 | 62.67 | 58.67 | 90.67 | 62.00 | 65.67 |
| Quadratic (Q) | 91.33 | 70.67 | 44.67 | 76.00 | 68.00 | 83.33 | 63.33 | 61.33 | 93.33 | 64.67 | 71.67 |
| DiagLinear (DL) | 79.33 | 60.00 | 21.33 | 55.33 | 60.67 | 58.67 | 54.00 | 43.33 | 76.67 | 50.67 | 56.00 |
| DiagQuadratic (DQ) | 85.33 | 57.33 | 20.67 | 50.67 | 55.33 | 58.67 | 54.00 | 50.67 | 88.67 | 53.33 | 57.47 |
| Mahalanobis (M) | 92.00 | 51.33 | 69.33 | 62.67 | 82.00 | 70.00 | 70.00 | 72.00 | 70.00 | 66.67 | 70.60 |
| L+D+DL+DQ+M majority | 86.67 | 66.00 | 47.33 | 68.00 | 75.33 | 78.00 | 66.00 | 62.67 | 91.33 | 62.00 | 70.33 |
| **L+D+M majority** | 91.33 | 62.67 | 52.00 | 72.67 | 76.67 | 81.33 | 68.00 | 66.00 | 92.67 | 65.33 | **72.87** |

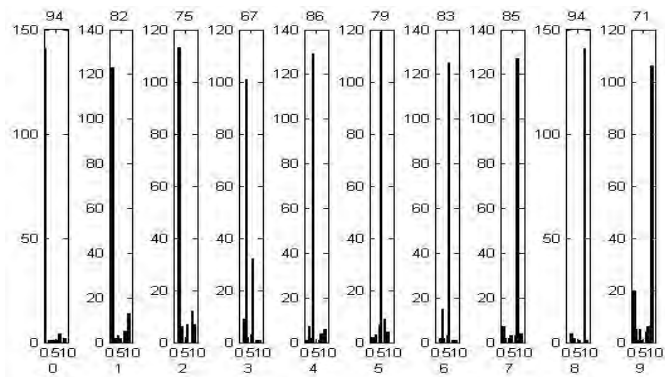

Figure 5: Classification result for training stage using combination classifier.

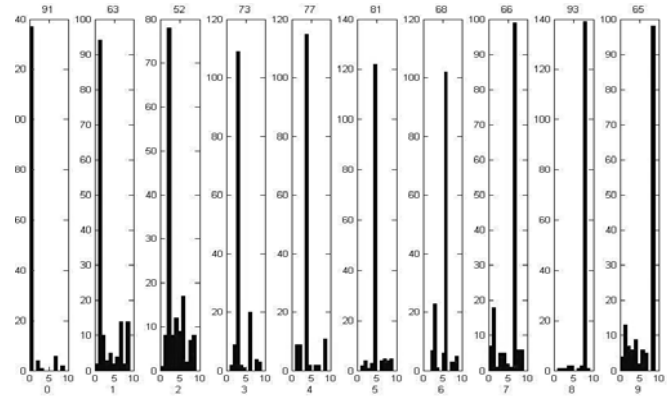

Figure 6: Classification result for testing stage using combination classifier.

## VIII. CONCLUSION

The results are satisfactory. But more accuracy needs to be achieved for actual implementation of the recognition system since mistake in one digit may have unforeseen consequences in understanding the whole number in any application. In future research, the features extracted will also be used with other classifiers like ANN and SVM to improve accuracy. The geometric and structural features used in this work are script independent. Some script dependent features may also be incorporated for improvement of results.

AUTHORS PROFILE

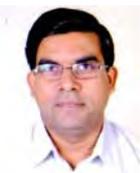

Vikas J. Dongre received B.E and M.E. degree in Electronics in 1991 and 1994 respectively. He served as lecturer in SSVPS engineering college Dhule (MS), India from 1992 to 1994, He Joined Government Polytechnic Nagpur in 1994 as Lecturer. He is presently working as lecturer in selection grade at in Government Polytechnic Gondia (MS). He is member of Indian Society for Technical Education (ISTE) and Institute of Electronics and Telecommunication (IETE). His areas of interests include Microcontrollers, embedded systems, pattern recognition, Education Technology. He has published six research papers in international journals and conferences. Presently he is pursuing for Ph. D. in Offline Devnagari Character Recognition from RTM Nagpur University, Nagpur (MS), India.

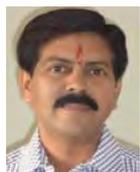

Vijay H. Mankar received M. Tech. degree in Electronics Engineering from VNIT, Nagpur University, India in 1995 and Ph. D. (Engg.) from Jadavpur University, Kolkata, India in 2009 respectively. He has more than 18 years of teaching experience and presently working as a Lecturer (Selection Grade) in Government Polytechnic, Nagpur (MS), India. He is member of Indian Society for Technical Education (ISTE) and Institute of Electronics and Telecommunication (IETE). He has published more than 40 research papers in international conferences and journals. His field of interest includes digital image processing, data hiding and watermarking.